%% file: main.tex
\documentclass[runningheads]{llncs}

\usepackage{eccv}

\usepackage{eccvabbrv}

\usepackage{graphicx}
\usepackage{booktabs}

\usepackage[accsupp]{axessibility}  

\usepackage[pagebackref,breaklinks,colorlinks,citecolor=eccvblue]{hyperref}

\usepackage{orcidlink}

\begin{document}

\title{\emph{StableDrag}: Stable Dragging for Point-based Image Editing} 


\author{Yutao Cui\inst{1,2} \and
Xiaotong Zhao\inst{2} \and
Guozhen Zhang\inst{1} \and
Shengming Cao\inst{2} \and
Kai Ma\inst{2} \and
Limin Wang\inst{1}
\thanks{Corresponding author, E-mail: lmwang@nju.edu.cn}
}

\authorrunning{Yutao Cui et al.}

\institute{
$^{1\;\;}$State Key Laboratory for Novel Software Technology,
Nanjing University \; \\
$^{2\;\;}$Tencent Inc. \\
\href{https://stabledrag.github.io/}{\email{https://stabledrag.github.io/}}
}

\maketitle

\input{sec/0_abstract} 

\begin{figure*}[t]
\centering
\includegraphics[width=0.98\linewidth]{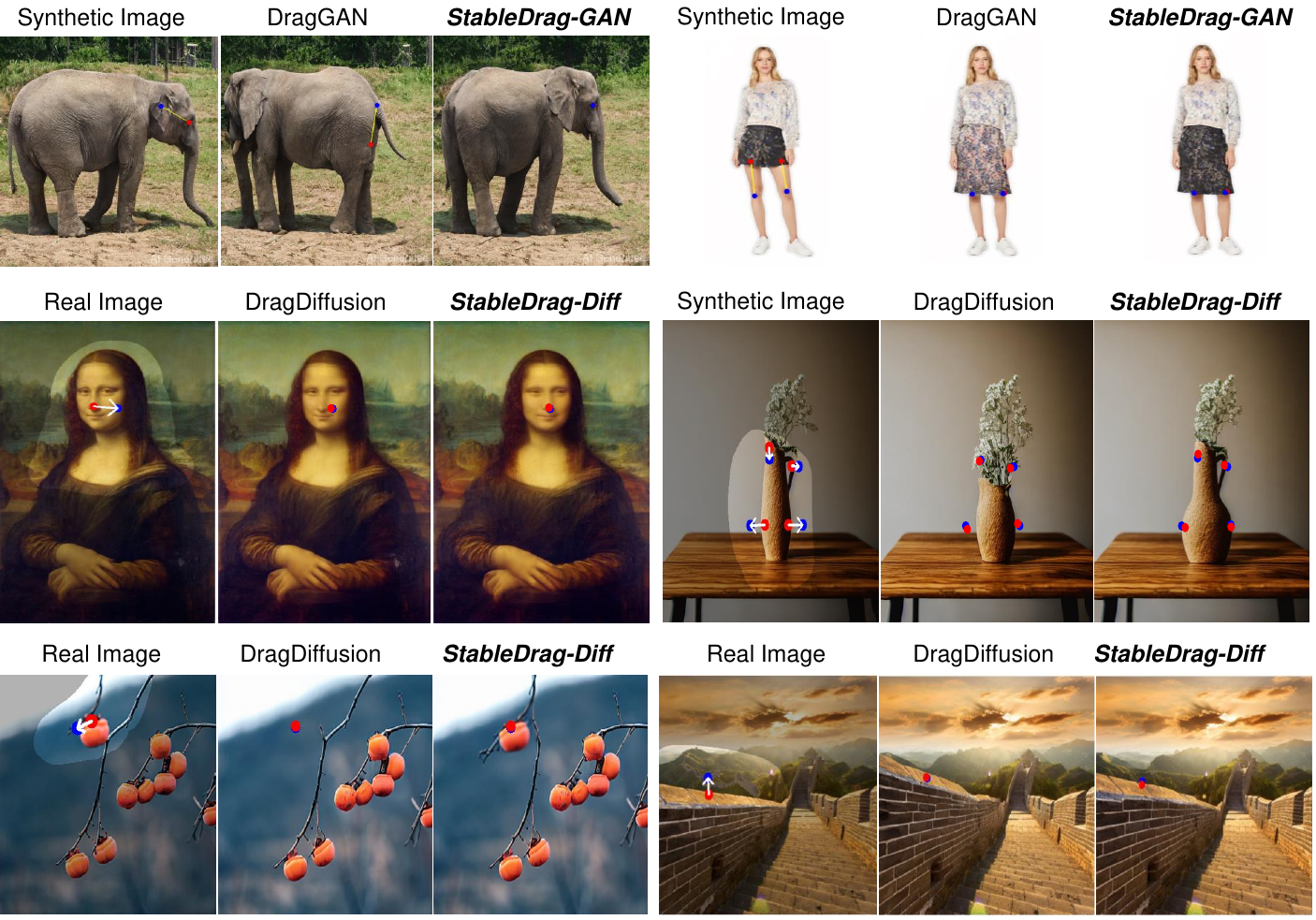}
\caption{\textbf{The comparison between DragGAN/DragDiffusion \cite{shi2023dragdiffusion} and our proposed StableDrag.} StableDrag-GAN and StableDrag-Diff are our proposed methods constructed upon GAN and Diffusion models respectively.  Given an image input (synthetic image by GAN/Diffusion model, or real image), users can assign handle points (\textcolor{red}{red} points) and target points (\textcolor{blue}{blue} points) to drive the semantic positions of the handle points to reach corresponding target points. The example of the Mona Lisa portrait and examples in the last row are the real-image inputs, while the others are synthetic from StyleGAN2 or Stable Diffusion-V1.5~\cite{rombach2022high} models. The examples demonstrate that our method achieves more precise point-level manipulation and generates higher-quality editing image than DragGAN and DragDiffusion.}
\vspace{-7mm}
\label{fig:figure-1}
\end{figure*}

\input{sec/1_intro}
\input{sec/related_work}
\input{sec/method}
\input{sec/experiments}

\input{sec/conclusion}

\clearpage  

\begin{center}
    \centering
    \captionsetup{type=figure}
    \includegraphics[width=\textwidth]{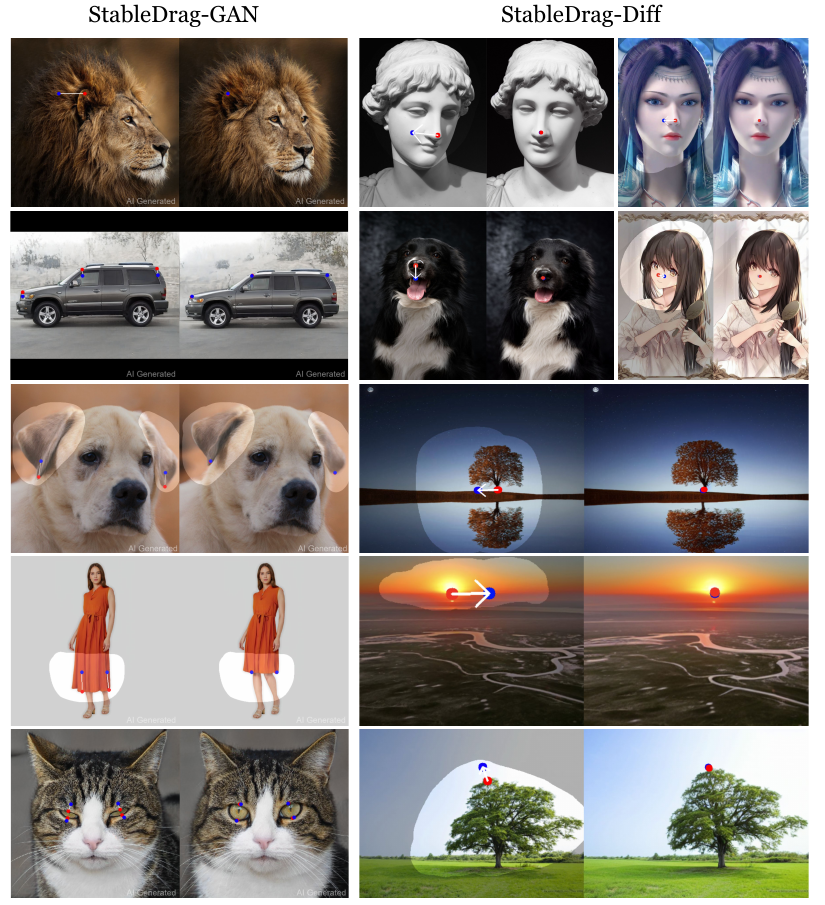}
    \vspace{-4mm}
    \captionof{figure}{ \textbf{More results of our StableDrag-GAN and StableDrag-Diff.}}
    \label{fig:supp_vis}
\end{center}

\section*{Appendix} We provide more visualization results of our StableDrag, including StableDrag-GAN and StableDrag-Diff, which are built upon DragGAN~\cite{pan2023drag} and DragDiffusion~\cite{shi2023dragdiffusion} respectively. It can be seen from the Fig.~\ref{fig:supp_vis}, our method can produce precise and stable dragging performance on a majority of scenarios. Furthermore, we provide more visualization results and give detailed comparison between the StableDrag and FreeDrag~\cite{ling2023freedrag} dragging process in \url{https://stabledrag.github.io/}. Code will be released upon acceptance.

%
%
\bibliographystyle{splncs04}
\bibliography{main}
\end{document}

%% file: sec/0_abstract.tex
\begin{abstract}
Point-based image editing has attracted remarkable attention since the emergence of DragGAN. 
Recently, DragDiffusion further pushes forward the generative quality via adapting this dragging technique to diffusion models.
Despite these great success, this dragging scheme exhibits two major drawbacks, namely inaccurate point tracking and incomplete motion supervision, which may result in unsatisfactory dragging outcomes.
To tackle these issues, we build a stable and precise drag-based editing framework, coined as \textbf{\emph{StableDrag}}, by designing a discirminative point tracking method and a confidence-based latent enhancement strategy for motion supervision.
The former allows us to precisely locate the updated handle points, thereby boosting the stability of long-range manipulation, while the latter is responsible for guaranteeing the optimized latent as high-quality as possible across all the manipulation steps.
Thanks to these unique designs, we instantiate two types of image editing models including StableDrag-GAN and StableDrag-Diff, which attains more stable dragging performance, through extensive qualitative experiments and quantitative assessment on DragBench.

\keywords{Stable dragging \and Image editing \and Drscriminative tracking \and Confident motion supervision}

\end{abstract}

%% file: sec/1_intro.tex
\section{Introduction}
\label{sec:intro}

Controllable image editing with generative models~\cite{kingma2013vae, mirza2014cgan, goodfellow2014gan, karras2019stylegan1, karras2020stylegan2, ling2023freedrag} has achieved remarkable achievements in the past few years, which can customize the generative results for further refinement purposes.
Recently the pioneering DragGAN~\cite{pan2023drag} has largely pushed forward accurate image editing with interactive point-based manipulation, that is, driving semantic objects based on user-input handle points toward the corresponding target points.
DragGAN formulates a novel dragging technique, primarily contains motion supervision and point tracking, where the former supervises the local patches around the handle points to move towards the target points step by step, while the latter is responsible for locating the updated handle points at each step.

Despite the great success of DragGAN, its editing ability is still constrained by the inherent model capacity and generality of generative adversarial networks.
Therefore, recent works~\cite{shi2023dragdiffusion,mou2023dragondiffusion} resort to diffusion models~\cite{ho2020ddpm, dhariwal2021beatgan, rombach2022ldm, ramesh2022dalle2, esser2021imagebart, gu2022vqdiffusion, nichol2021glide, saharia2022imagen, gal2022textualinversion, kumari2022customdiffusion, kawar2022imagic} for high-quality drag-style image editing.
A representative work DragDiffusion~\cite{shi2023dragdiffusion} explores to adapt the dragging scheme to diffusion models, i.e., first fine-tuning a LoRA, then optimizing the latent at a single diffusion step, finally denoising the optimized latent based on MasaCtrl~\cite{cao2023masactrl}.
For the key component of diffusion latent optimization, it directly follows the DragGAN's convention of iteratively conducting motion supervision and point tracking. 
We analyze that the current dragging scheme still suffers from the following issues.
\romannumeral1) \emph{\textbf{Inaccurate point tracking.}} These methods leverage the feature difference as the similarity measurement to track the updated handle points, which is insufficient to precisely locate the right ones from the distractors (i.e., the around misleading points with similar content).
Especially in diffusion models, since the features are sampled from the intermediate diffusion process with much noise injection, the updated points become increasingly challenging to be distinguished from their local surroundings.
This may lead to unsatisfactory dragging outcomes, as showcased by the examples of the Mona Lisa portrait and the vase in Fig.~\ref{fig:figure-1}.
\romannumeral2) \emph{\textbf{Incomplete motion supervision.}} 
During the motion supervision process, the latent may not be adequately optimized at certain steps, resulting in a deterioration of the manipulation quality (see examples of the elephant and the woman in Fig.~\ref{fig:figure-1}) as well as the point tracking drift.
In diffusion models, the latent is more stable and harder to manipulate than GAN's~\cite{shi2023dragdiffusion}, especially when fine-tuning the LoRA on a specific image, which may aggravate the problem.

Considering the aforementioned issues, we argue that there are two primary principles for designing a more stable dragging framework.
First, \emph{a robust yet efficient point tracking} method is required, to avoid locating the incorrect points and increasing much latency, thus enabling the point-based drag to be precise.
Second, we should guarantee \emph{the motion supervision to be complete} at each optimization step, so as to keep the editing content as high-quality as possible across all the manipulation process, and fully unleash the strong restoring power of generative models. In addition, complete motion supervision can enhance the similarity between the content of the given handle points and the updated points, preventing the accumulation of tracking errors.

Driven by the above analysis, we re-formulate the dragging scheme of point tracking and motion supervision in DragGAN and DragDiffusion, and present a more stable dragging framework for point-based image editing, coined as~\textbf{StableDrag}.
Specifically, inspired by the success in visual object tracking~\cite{siamfc,atom,mixformer}, we try to derive a simple yet powerful point tracking model, in the form of a convolution filter, from a discriminative learning loss.
This model is capable of suppressing the tracking confidence score of the distractor points as well as enhancing that of the handle points.
At the beginning of the manipulation steps, we update the tracking model weights under the supervision of a tailored similarity learning function.
Once the tracking model is prepared, we employ it, in conjunction with the original feature difference method for robust and precise point tracking.
Notably, this approach scarcely increases inference latency, since we only need to optimize the simple tracking model (i.e., a single convolution filter) at the initial manipulation step.
Furthermore, we design a confidence-based latent enhancement strategy, to make motion supervision complete enough at each step.
In detail, we utilize the tracking confidence score of the handle points to assess the quality of the current manipulation process. 
Normally, we use the same manner of motion supervision as DragDiffusion.
Nevertheless, when the quality score falls below an acceptable threshold, we employ the template features (i.e., the initial features of the given start handle points) to supervise that of the current handle points' content, until its confidence score is satisfactory.
Thanks to the unique designs for dragging scheme, we instantiate two types of image editing models including StableDrag-GAN and StableDrag-Diff, built on GAN and Diffusion models respectively, which attains more stable and precise drag performance.
Our contributions are summarized as follows:
\begin{itemize}
\item[$\bullet$] We propose a discriminative point tracking method, which allows the model to accurately distinguish the updated handle points from the distractor ones, hence promoting the stability of dragging.
\item[$\bullet$] We devise a confidence-based latent enhancement strategy for motion supervision, which can improve the optimization quality at each manipulation step.
\item[$\bullet$] Under these designs, we build \textbf{StableDrag}, a point-based image editing framework, upon different generative models including GAN and Stable Diffusion. Through extensive qualitative experiments on a variety of examples and quantitative assessment on DragBench~\cite{shi2023dragdiffusion}, we demonstrate the effectiveness of our StableDrag-GAN and StableDrag-Diff. 
\end{itemize}

%% file: sec/related_work.tex
\section{Related Work}
\label{sec:relate}

\subsection{Image Editing}
Image editing is a hot topic with a wide range of applications.
Generative Adversarial Networks (GANs) have made significant strides in the field of image generation\cite{NIPS2014_5ca3e9b1,karras2019style}, leading to numerous prior image editing techniques\cite{endo2022user,pan2023drag,abdal2021styleflow,leimkuhler2021freestylegan,patashnik2021styleclip,zhu2016generative} being founded upon the GAN framework. Nonetheless, the model capacity of GANs remains somewhat constrained, as well as the challenge of effectively transforming real images into GAN latent spaces\cite{abdal2019image2stylegan,creswell2018inverting,lipton2017precise}, the practicality of these approaches was inevitably constrained.
Recently, large-scale text-to-image diffusion models have produced remarkably realistic generation results~\cite{ho2020ddpm, dhariwal2021beatgan, rombach2022ldm, ramesh2022dalle2, esser2021imagebart, gu2022vqdiffusion, nichol2021glide, saharia2022imagen, gal2022textualinversion, kumari2022customdiffusion, kawar2022imagic}, which have given rise to numerous diffusion-based image editing methods\cite{hertz2022prompt,cao2023masactrl,mao2023guided,kawar2023imagic,parmar2023zero,liew2022magicmix,mou2023dragondiffusion,tumanyan2023plug,brooks2023instructpix2pix,meng2021sdedit,bar2022text2live,epstein2023selfguidance}. These techniques primarily strive to edit images by adjusting the prompts associated with the image. Nevertheless, as many editing endeavors prove challenging to convey through text, the prompt-based strategy frequently modifies the image's high-level semantics or styles, thereby lacking the capability to achieve precise pixel-level spatial manipulation.

In order to facilitate fine-grained editing, a number of studies have been proposed to execute point-based modifications, such as \cite{pan2023drag,endo2022user,wang2022rewriting}. In particular, DragGAN has exhibited remarkable dragging-based manipulation through two straightforward components: the optimization of latent codes to shift the handle points towards their desired destination points and a point tracking mechanism to locate the updated handle points. However, its generality is constrained due to the limited capacity of GAN. 
DragDiffusion~\cite{shi2023dragdiffusion} and DragonDiffusion~\cite{mou2023dragondiffusion} further extend the dragging scheme to diffusion models to leverage its excellent generative capacity.
FreeDrag \cite{ling2023freedrag} has proposed to improve DragGAN by introducing a point-tracking-free paradigm.
In this work, we explore a new dragging scheme with re-formulating a confident motion supervision module and a discriminative point tracking module, enabling stable point-based image editing.

\subsection{Visual Tracking}
Since the proposed discriminative point tracking takes inspiration from the visual tracking research, we give a brief overview for these methods.
We divide the works into three categories.
First, correlation-filter-based trackers~\cite{mosse,kcf,atom} learned an online target-dependent discriminative model for tracking. 
\cite{mosse,kcf} employed online correlation filters to distinguish targets from background and obtains good performance with a high running speed, which is very practical until now.
Second, Siamese-based trackers~\cite{siamfc,siamrpn} attract a lot of attention due to its simplicity and efficiency.
These methods combined a correlation operation with the Siamese network, modeling the appearance similarity and correlation between the target and search. 
SiamFC~\cite{siamfc} employed a Siamese network to measure the similarity between the template and the search area with a high tracking speed. SiamRPN++~\cite{siamrpnPlus} improved cross correlation to depth-wise cross correlation, which can increase both the performance and efficiency.
Finally, some recent trackers~\cite{mixformer,transt,mixformerv2} introduced a transformer-based integration module to capture the similarity between the target and search region. Inspired by these findings, we devise a robust point tracking model via discriminative learning.
Different from these works, we build the tracking model on top of the intermediate feature of GAN or diffusion models to leverage their discriminativeness and only optimize the tracking model, which is effective yet efficient.

%% file: sec/method.tex
\section{Method}
\label{sec:method}

\subsection{Preliminary on Point-based Dragging}
Firstly, we briefly review the recent literature on the point-based dragging framework behind GAN and diffusion models, which are the basics of our work.

\paragraph{\textbf{DragGAN.}} 
Given an image generated by GAN models~\cite{karras2020stylegan2,karras2021stylegan3}, in conjunction with the user-input handle points $\{{p_{i}=(x_{pi}, y_{pi}), i=1,2,...,n}\}$ and the target points $\{{t_{i}=(x_{ti}, y_{ti}), i=1,2,...,n}\}$, DragGAN aims to drive the content at every handle point $p_i$ move towards their corresponding target point ${t_i}$.
In this sense, the primary concern lies in how to precisely control the point-level editing while maintaining high image fidelity.
To achieve the goal, DragGAN tailors a novel paradigm, which involves repeated motion supervision and point tracking.
Considering the generator's characteristic that the intermediate features are very discriminative, they leverage a simple online motion supervision loss to optimize the latent code.
When denoting the local region around $p_i$ as $\Theta(p_i)$, i.e, the pixels whose distance to $p_i$ is less than the radius $r_i$, the loss can be defined as:
\begin{equation}
    \mathcal{L}_{1} = \sum_{i=0}^{n} \| \mathbf{F}(\Theta(p_i)) - \mathbf{F}(\Theta(p_i+d_i) \|_1 + \eta \| (\mathbf{F} - \mathbf{F}^0) \cdot (1-\mathbf{M}) \|_1,
    \label{eq:motion_1}
\end{equation}
where $\mathbf{F}$ represents for the intermediate feature at current optimization step, $\mathbf{F}^0$ is the feature at initial step, $n$ is the number of handle points, $d_i = \frac{t_i - p_i}{\|t_i - p_i\|_2}$ is a deviation vector and $\mathbf{M}$ is the pre-defined mask to control the changing area.
Particularly, since the $\mathbf{F}(\Theta(p_i))$ gets detached, the content of current $p_i$ will be motivated to $t_i$ by a small step.
However, due to the inherent indeterminacy of optimization, it is hard to guarantee $p_i$ to approach $p_i+d_i$. Consequently, they utilize a simple feature difference method as point tracking to determine the updated state of $pi$.
The above optimization process iterates until each of the handle points $p_i$ converges to their respective target points $t_i$.

\paragraph{\textbf{DragDiffusion.}}
DragDiffusion~\cite{shi2023dragdiffusion} extends the point-based editing framework to diffusion models, such as Stable Diffusion (SD-V1.5~\cite{rombach2022high}), so as to unleash its strong power of high stability and superior generation quality.
This editing method involves three sub-processes, i.e., finetuning a LoRA on the real image, optimizing the latent on a certain diffusion step and denoising the updated latent to generate the edited image.
Specifically, they adopt the same dragging formulation of repeated motion supervision and point tracking on a single intermediate diffusion step to manipulate the latent.
Besides, a LoRA finetuing strategy is employed to preserve the image identity through the whole manipulation process.
Finally, a self-attention control mechanism MasaCtrl~\cite{cao2023masactrl} is used to enhance the consistency between the original image and the edited image.

\begin{figure*}[t]
\centering
\includegraphics[width=0.95\linewidth]{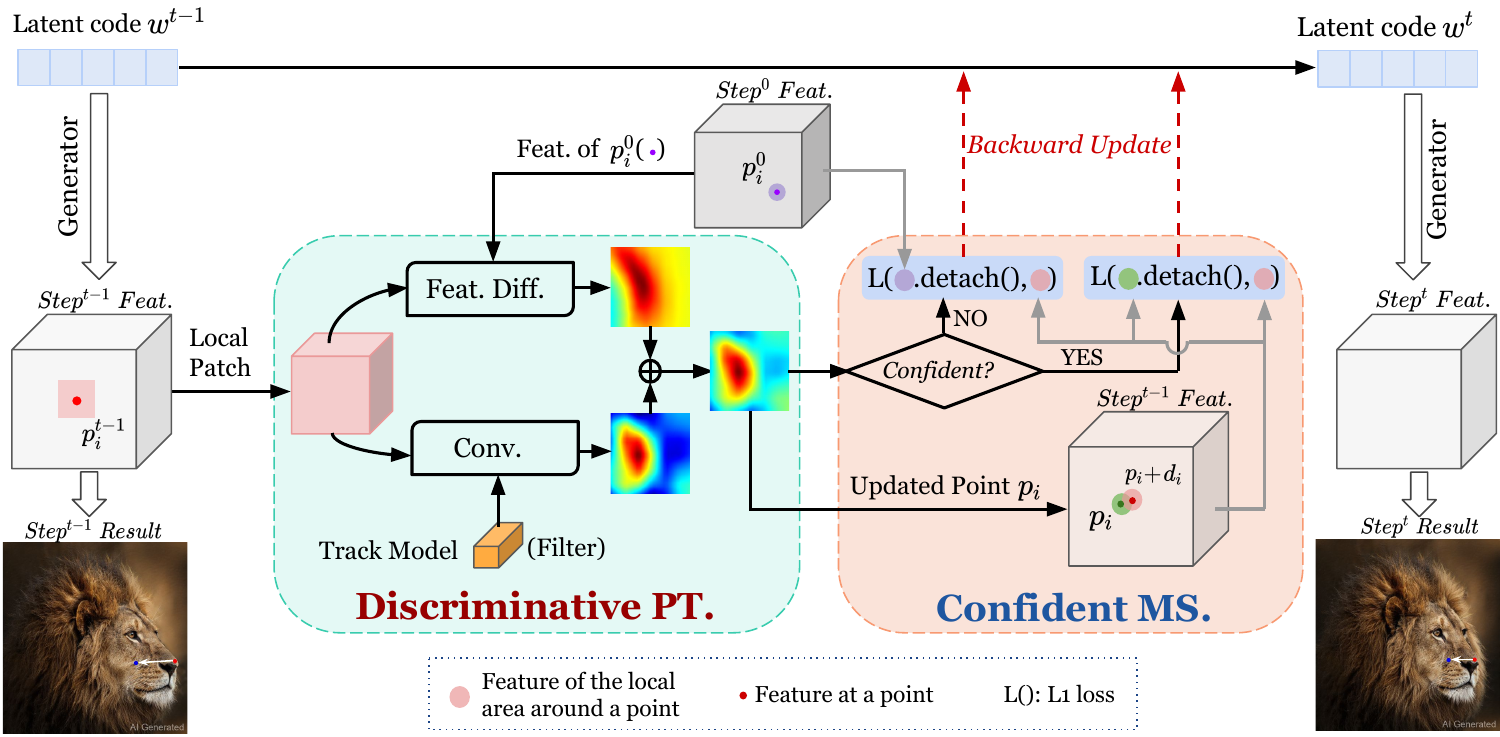}
\caption{{\bf Illustration of our dragging scheme for an intermediate single-step optimization.} The core of the dragging pipeline illustrated herein is based on GAN, whereas the one based on diffusion models remains the same. `Discriminative PT.' denotes for discriminative point tracking module and `Confident MS.' represents for confident motion supervision process. $P_{i}$ means the current handle point at $i^{th}$ step optimization. Notably, the tracking model, in the form of a convolution filter, is only learned at the first optimization step and can be just employed in the subsequent steps. Details about its learning process at the first step are described in Fig.~\ref{fig:track_model}. The latent code $w$ is supposed to be optimized via the backward updating across all steps.}
\vspace{-4mm}
\label{fig:framework}
\end{figure*}

\subsection{Overview}
As illustrated in Fig.~\ref{fig:figure-1}, DragGAN and DragDiffusion may result in deteriorated editing images due to the imprecise point tracking and incomplete motion supervision.
Therefore, in this work, we cast attention on the current dragging technique to achieve more stable and precise image manipulation.
The developed dragging pipeline is illustrated in Fig.~\ref{fig:framework}, which comprises a discriminative point tracking module and a confident motion supervision module.
Specifically, we design a new point tracking approach that integrates the original feature difference with the tracking score yielded from a learned discriminative tracking model, thereby boosting the point tracking accuracy as well as the drag precision.
Based on the tracking score, we then explore a confidence-based latent enhancement strategy to achieve complete enough motion supervision.
We also observe that DragGAN masters large deformation and creative content (e.g., transforming a lion with its mouth closed into a roaring state) within a short run-time.
While DragDiffusion is good at generating superior-quality and higher-fidelity editing outcomes.
To enable the dragging model to accommodate a wide range of scenarios, we build StableDrag upon both DragGAN and DragDiffusion with the designed dragging scheme. 
In this section, we will introduce the proposed dragging method in details.

\subsection{Discriminative Point Tracking}
Point tracking serves as a pivotal function in identifying the updated handle points $p_i$, to circumvent dragging erroneous points and produce unsatisfactory editing results.
The prevalent approach employed in DragGAN and DragDiffusion is straightforward, that is, conducting nearest neighbor search by identifying the position with minimal feature difference to the initial feature template of $p_0$.
However, this entirely ignores background appearance information, which is crucial for discriminating the handle points from the similar ones in the complex scene.
Particularly, in diffusion models, since the supervision features are extracted from the intermediate diffusion stage, which incorporates substantial noise, it becomes progressively difficult to discern the updated points.
For instance, as shown in the case of the Mona Lisa portrait of Fig.~\ref{fig:figure-1}, the handle point of the nose possesses similar appearance with the adjacent points, which causes the misleading location in DragDiffusion.
Therefore, in this work, we explore an alternative method for accomplishing more discriminative yet simple point tracking.

Distinguishing the given handle points from the distractors can be addressed using a learnable discriminative tracking model.
In our design, the point tracking model constitutes the weights of a convolutional layer, providing the point classification scores as output.
In detail, we propose to learn a function $g(\mathbf{F}(\Theta_2), z_i)$, where $g$ denotes a convolution function, $\Theta_2$ is the local patch around the current handle point $pi$ and $z_i$ is the learned tracking model, which returns a high score if the tracking model $z_i$ matches the content at a certain position and discerns it as the updated handle point $p_i$, and a low score otherwise.
In particular, the tracking model $z_i$ is learned before the latent optimization and keep unchanged across all the manipulation steps.
In this sense, this approach scarcely increases the editing runtime.
Finally, we merge the classification score yielded by the tracking model with the original feature difference score, so as to achieve both discriminative and precise point location.
The detailed procedure of the discriminative point tracking is illustrated in Fig.~\ref{fig:framework}.

\begin{figure}[t]
\centering
\includegraphics[width=0.95\linewidth]{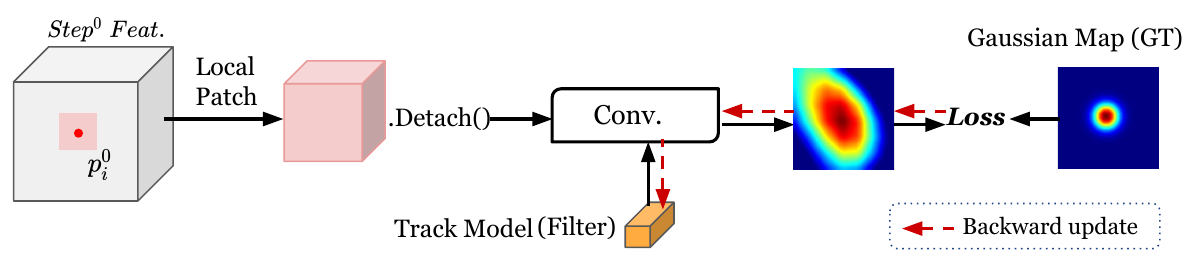}
\caption{{\bf Learning process of our point tracking model.} It is only performed before the manipulation process. The initial feature of the local patch gets detached, indicating that only the tracking model is supposed to be optimized. The tracking model weight is initialized with the the template feature $f_i$.}
\vspace{-3mm}
\label{fig:track_model}
\end{figure}

Formally, given the local patch $\Theta_2 (p_i, r_2)=\{(x,y)~\big|~\lvert x-x_{pi}<r_2 \rvert,\lvert y-y_{pi}<r_2 \rvert\}$, the tracked point $p_i$ is updated as:
\begin{equation}
\begin{aligned}
    & S(\Theta_2) = \lambda * e^{-\| \mathbf{F}(\Theta_2) - f_i \|_1} + (1-\lambda)* g(\mathbf{F}(\Theta_2), z_i), \\
    & p_i := \mathop{\arg\max}_{q_i \in \Theta_2 (p_i, r_2)} S(\Theta_2(p_i, r_2)), \\
    & s_i = \mathop{\max}_{q_i \in \Theta_2 (p_i, r_2)} S(\Theta_2(p_i, r_2)),
    \label{eq:tracking}
\end{aligned}
\end{equation}
where $S(\Theta_2)$ represents for the tracking confidence score map of the local patch $\Theta_2$, $\lambda$ is the weighting factor, $f_i=\mathbf{F}^0(p_i^0)$ is the original feature of the initial handle point $p_i^0$ at the step-0, and $s_i$ is the maximal tracking confidence score at the current step, which is used to guide the motion supervision. 
In the terms of $S(\Theta_2)$, the former one measures the feature difference the template and the search region. 
Although it can provide accurate point localization in the majority of instances, it may be misled by the distractor points.
Therefore, the second term is responsible to improve the tracking robustness with the discriminative learning, i.e., suppressing the score of surrounding points during the initial optimization process for $z_i$.
Unlike the plain feature difference method, this tracking model is capable of leveraging background information and harnessing distinguishing characteristics of the intermediate feature, thus providing a valuable enhancement to the original approach.

\paragraph{\textbf{Learning for tracking model $\mathbf{z_i}$.}} The learning of the point tracking model $z_i$, which is a convolutional filter with the size of $1\times C\times 1\times 1$, is performed before the manipulation process.
Overview of the learning process is shown in Fig.~\ref{fig:track_model}.
We use $f_i$ to initialize $z_i$ and update the weights under the supervision of the following loss:
\begin{equation}
    \mathcal{L}_{track} = \| g(\mathbf{F_0}(\Theta_2(p_i, r_2)), z_i) - y_i \|^2.
\end{equation}
Here, $\mathbf{F_0}$ denotes the initial feature at step-0, $y_i$ represents for the ground-truth label, which is the desired confidence scores at each position, generally set to a Gaussian function centered at $p_i$.
During the learning process, the gradient is not back-propagated through $\mathbf{F_0}(\Theta_2(p_i, r_2))$.
In other words, we only need to optimize the tracking model $z_i$, allowing for rapid convergence.
Through the optimization, we highlight the handle points while simultaneously suppressing the confidence score of the background points.
Then in the subsequent manipulation steps, the tracking model $z_i$ keeps unchanged for efficiency.

\subsection{Confident Motion Supervision}

Motion supervision is the core to progressively encourage the points to move towards their intended destination.
DragGAN employs an online loss in equation~(\ref{eq:motion_1}) to achieve the goal, however may yielding unsatisfactory results in long-range drag.
Alternatively, we devise a confident motion supervision component based on the tenet that,
\emph{not only ensuring high-quality and comprehensive supervision at each step but also allowing for suitable modifications to accommodate the novel content creation for the updated states}.
For example, the case of a woman wearing a skirt in Fig.~\ref{fig:figure-1} demonstrates the significance of complete supervision in maintaining visual coherence.

To attain the above goal, we propose a confidence-based latent enhancement strategy as shown in Fig.~\ref{fig:framework}.
Firstly, we introduce the maximal value of the tracking score, i.e. $s_i$, to represent the current supervision confidence, and the confidence score $s_1$ at the step-1 to produce the threshold for enhancement strategy. 
Normally, the original motion supervision as in equation~(\ref{eq:motion_1}) is employed when we discern the current state being confident enough.
If the current confidence score falls below the pre-defined threshold, we resort to the initial template for supervision.
The concrete enhancement supervision is defined as:
\begin{equation}
\begin{aligned}
     \mathcal{L}_{2} = & \sum_{i=0}^{n} \| \mathbf{F}^0(\Theta(p_i^0)) - \mathbf{F}(\Theta(p_i+d_i) \|_1 \\
    & + \eta \| (\mathbf{F} - \mathbf{F}^0) \cdot (1-\mathbf{M}) \|_1, 
\end{aligned}
\end{equation}
where $\mathbf{F}^0(\Theta(p_i^0))$ is the fixed template with no gradient back-propagating, which can enforce the content of updated points to mimic the initial state.
Moreover, the choice of whether to use this latent enhancement supervision is determined according to the following guidelines:
\begin{equation}
\mathcal{L}_{motion} =
\left\{
             \begin{array}{lr}
             \mathcal{L}_{1}, &  s_i > \tau * s_1, \\
             \mathcal{L}_{2}, & s_i <= \tau * s_1,
             \end{array}
\right.
\end{equation}
where $\tau$ is a threshold rate to control the enhancement strength.
In this way, we can prevent the current content of handle points from significantly deviate from the original template, thus achieving confident motion supervision.
On the other hand, when the confidence score surpasses the threshold, we rely on the dynamic motion supervision $\mathcal{L}_1$ to sustain a high editability.

\begin{figure}[t]
\centering
\includegraphics[width=\linewidth]{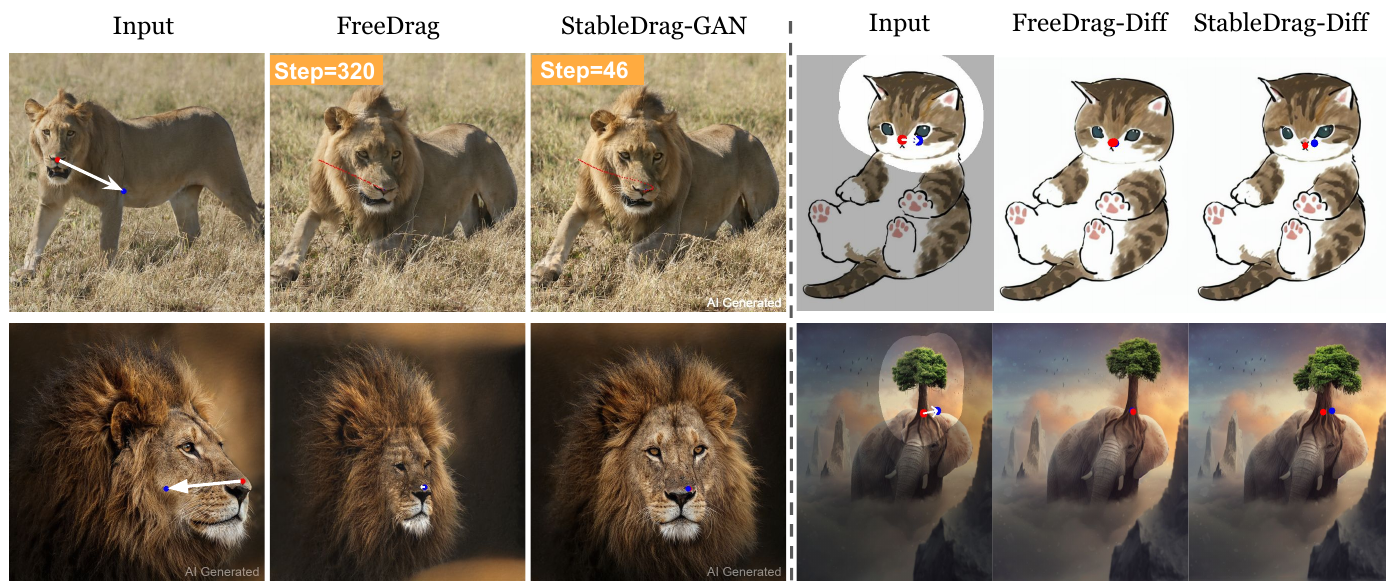}
\caption{{\bf Comparison between FreeDrag~\cite{ling2023freedrag} and our StableDrag.} For the example in the top left, handle points at each optimization step are visualized to show the difference of the optimization path of FreeDrag and our StableDrag-GAN. The example in the bottom left is to demonstrate our method's strength in creating novel content. And the others are to show that StableDrag can generate more precise dragging outcomes.}
\vspace{-3mm}
\label{fig:freedrag_comp}
\end{figure}

\paragraph{\textbf{Discussion.}}
To better expound the insight of the confident motion supervision, we make a comparison with the method proposed in FreeDrag~\cite{ling2023freedrag}, which employs an adaptive template and a linear search to set free the point tracking module.
First, The preset linear search in FreeDrag may impose restrictions on the flexibility of the latent optimization, thereby significantly increasing the difficulty of dragging.
As shown in the top-left example of Fig.~\ref{fig:freedrag_comp}, The handle points of FreeDrag frequently oscillate along the predefined path and necessitate 320 steps of optimization.
However, our method allows the handle points to move towards the destination along \emph{a more optimal path}, which is not linear, in only 46 steps.
Besides, FreeDrag struggles in generating creative and out-of-distribution content, as demonstrated by the bottom-left example in Fig.4, since it primarily relies on a template feature for supervision, even though an updating strategy is employed.
In contrast, our StableDrag-GAN can generate satisfactory creative content given a long-range dragging path, demonstrating better editability.

%% file: sec/experiments.tex
\section{Experiments}
\label{sec:exp}

\subsection{Implementation Details}
We implement the approach, including StableDrag-GAN and StableDrag-Diff, based on PyTorch.
During the process of optimizing the latent code $w_i$, we use Adam optimizer with learning rate of 0.01 for StableDrag-Diff and the 0.001 for StableDrag-GAN, which follows their default settings. In most cases, the hyper-parameters of $\lambda$ and $\tau$ are set to 0.3 and 0.4, respectively. For other parameters and model settings, we follow the default ones in DragGAN and DragDiffusion. The experiments are conducted on an NVIDIA V100 GPU.

\subsection{Qualitative Comparison}
\begin{figure*}[pt]
\centering
\includegraphics[width=\linewidth]{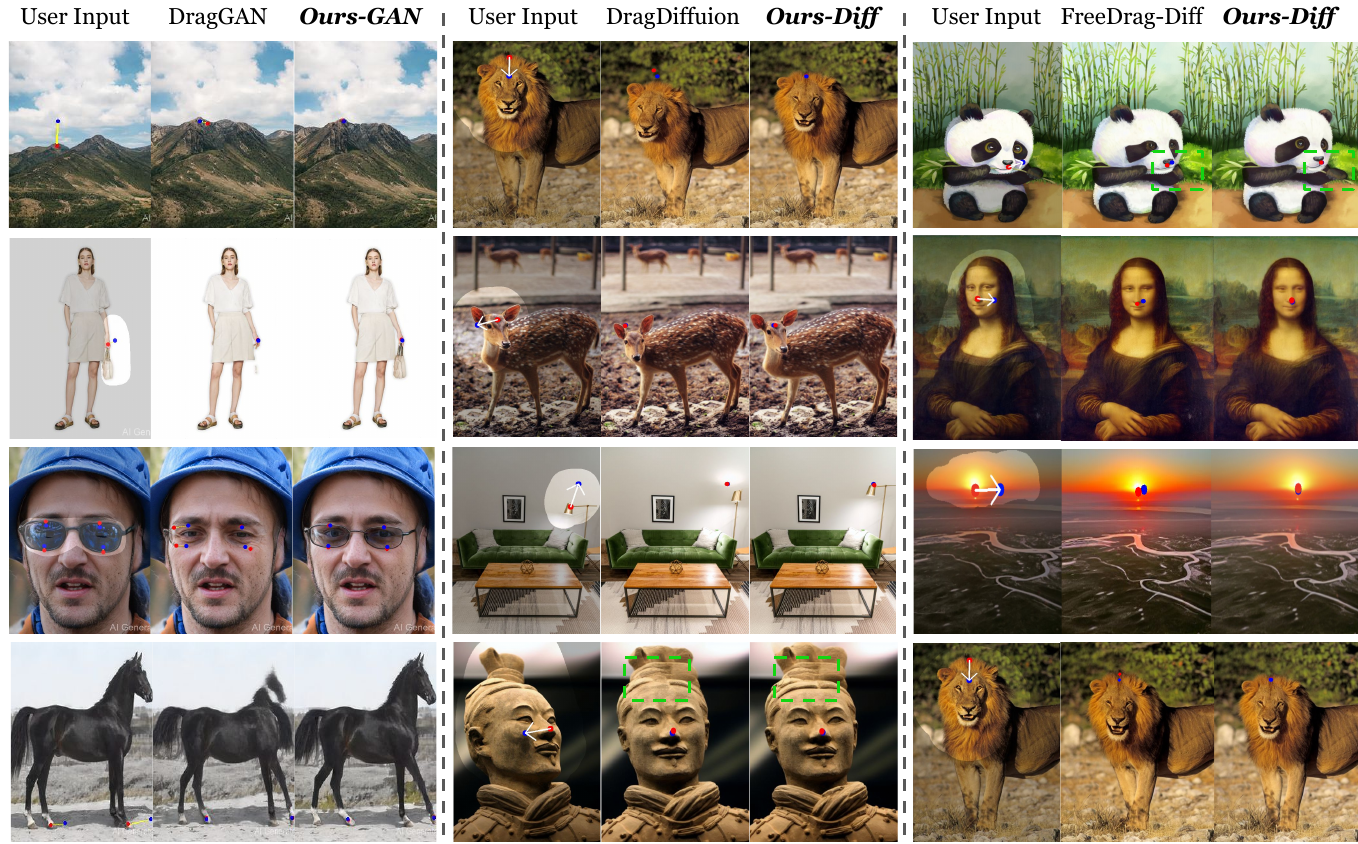}
\vspace{-3mm}
\caption{{\bf Comparison between DragGAN~\cite{pan2023drag}/DragDiffusion~\cite{shi2023dragdiffusion}/FreeDrag~\cite{ling2023freedrag} and our StableDrag.} As in DragGAN, users can optionally
draw a mask of the flexible region (\textcolor{gray}{brighter} area), keeping the rest of the image fixed. The green dashed box in the examples of \textit{the Terra Cotta Warriors Sculpture} and \textit{the Panda} is to show the differences in detail. Best viewed with zooming in.}
\vspace{-3mm}
\label{fig:qualit_reuslts}
\end{figure*}

Fig.~\ref{fig:qualit_reuslts} shows the qualitative results between DragGAN and StableDrag-GAN, DragDiffusion and StableDrag-Diff, FreeDrag-Diff and StableDrag-Diff for fair comparison. To evaluate the method's generality, for the GAN-based models, the input images are generated from StyleGAN2~\cite{karras2020stylegan2}. While for the Diffusion-based models, we input real images and use DDIM inversion to reconstruct them.
It can be seen that our method can more precisely move the handle points to the target points, such as the mountain peak, the lion's chin, the deer's forehead and the little lamp.
Besides, our StableDrag can generate higher-quality and higher-fidelity editing results, for example, maintaining the appearance of the bag, the glasses, the horse and the Terra Cotta Warriors sculpture.
We also compare our StableDrag-Diff with the FreeDrag~\cite{ling2023freedrag} based on Diffusion model. We can see that ours-Diff produces more precise results and maintains the details of the initial images.
This demonstrates the effectiveness of the proposed discriminative point tracking and confident motion supervision, which can achieve more stable dragging performance.

\begin{table}[pt]
\caption{Quantitative comparison on DragBench. `MD' denotes Mean Distance $\downarrow$ and `IF' is the Image Fidelity (1-LIPIPS) $\uparrow$.}
    \centering
    \fontsize{8pt}{5mm}\selectfont
    \setlength{\tabcolsep}{1.2mm}{
    \vspace{-2mm}
    \begin{tabular}{lcccccc}
        \toprule
        Optimization Steps & 60 & 80 & 100 \\
        Metric & MD/IF & MD/IF & MD/IF \\
        \hline
        DragDiffusion & 39.58/0.876 & 37.98/0.868 & 38.86/0.863 \\
        StableDrag-Diff & \textbf{36.36}/\textbf{0.893} & \textbf{36.98}/\textbf{0.884} & \textbf{35.92}/\textbf{0.869}  \\
        \bottomrule
    \end{tabular}
    }
    \label{tab1}
\end{table}

\subsection{Quantitative Results}
We quantitatively evaluate our method on DragBench~\cite{shi2023dragdiffusion}, comprising 205 samples with pre-defined drag points and mask.
We notice that, in DragBench, there are many examples that are not compatible with proper StyleGAN2 models, so we only conduct the experiments on DragDiffusion and ours-Diff.
We compare our StableDrag-Diff to DragDiffusion and use the same LoRA weights and the common hyper-parameters for fair comparison.
As shown in Table~\ref{tab1}, under three different setting of the optimization steps, StableDrag-Diff consistently outperforms the DragDiffusion, especially surpassing the baseline by 3.22 of Mean Distance score and 0.017 of Image Fidelity score with 60-step optimization.
This further indicates that our StableDrag can achieve promising results in editing accuracy and content consistency via the proposed confident motion supervision and discriminative point tracking.

\begin{figure*}[pt]
\centering
\begin{minipage}[t]{0.6\textwidth}
\centering
\includegraphics[width=\textwidth]{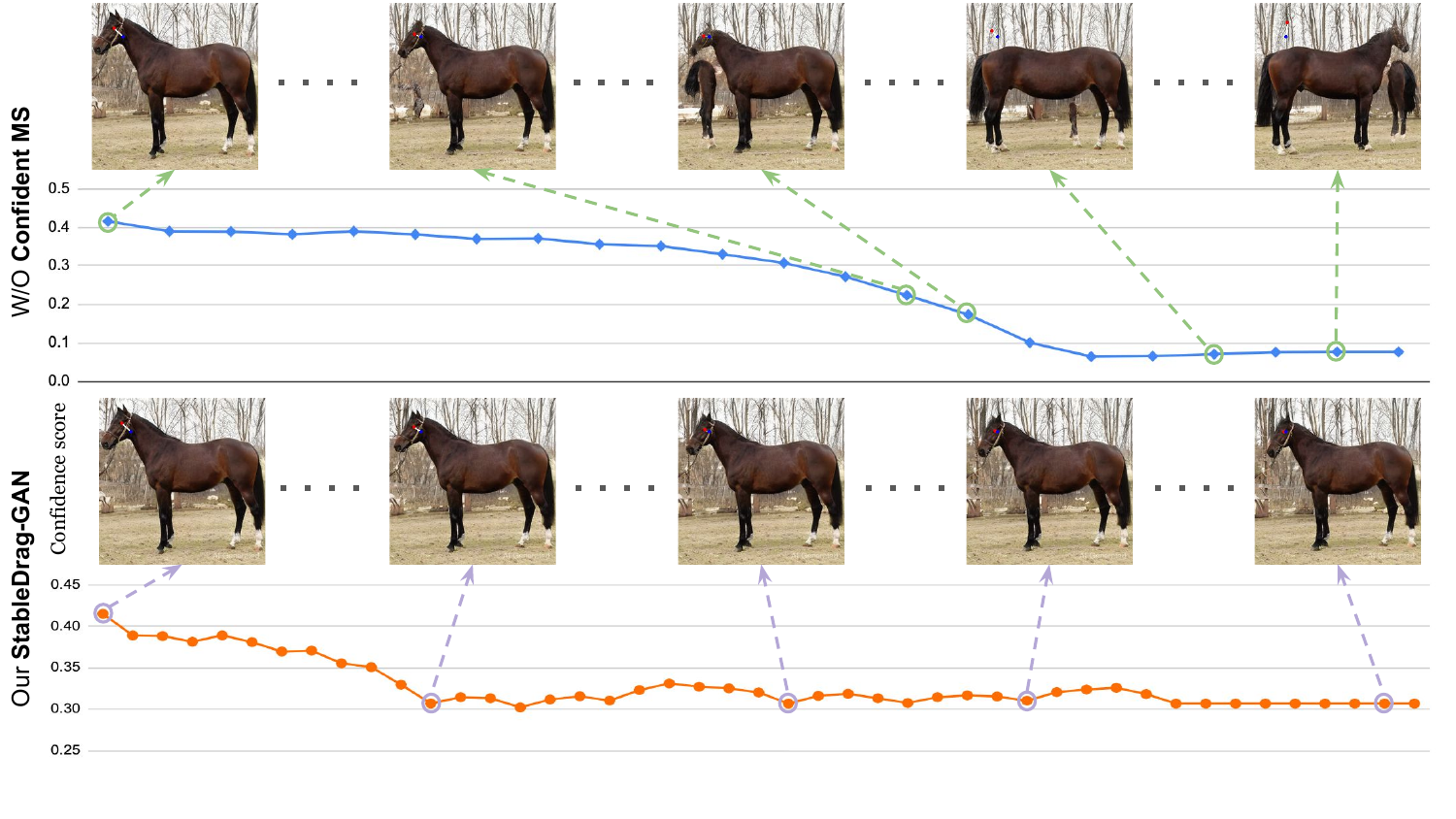}
\vspace{-4mm}
\caption{Effects of the latent enhancement strategy employed in confident motion supervision component. `W/O' denotes only using the original motion supervision method as in DragGAN.}
\vspace{-5mm}
\label{fig:ablation1}
\end{minipage}
\hspace{2mm}
\begin{minipage}[t]{0.36\textwidth}
\centering
\includegraphics[width=\textwidth]{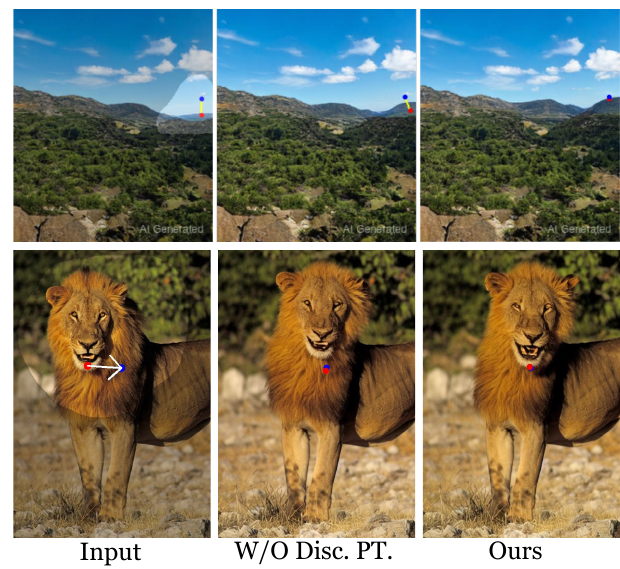}
\vspace{-4mm}
\caption{Effects of the discriminative point tracking. `W/O' denotes only using the plain feature difference method.}
\vspace{-5mm}
\label{fig:ablation2}
\end{minipage}
\end{figure*}

\subsection{Exploration Study}
To verify the effectiveness and give a thorough analysis on our proposed method, we perform a detailed ablation study through qualitative visualization based on both GAN and diffusion models, and quantitative evaluation on DragBench based on diffusion models.

\begin{table}[pt]
\caption{Effects of our discriminative point tracking and confident motion supervision. `DPT' denotes the discirminative point tracking and `CMS' is the confident motion supervision.}
    \centering
    \fontsize{8pt}{4.5mm}\selectfont
    \setlength{\tabcolsep}{1.2mm}{
    \vspace{-2mm}
    \begin{tabular}{lcc}
        \toprule
        Metric & Mean Distance $\downarrow$ & Image Fidelity $\uparrow$ \\
        \hline
        DragDiffusion & 39.58 & 0.876  \\
        StableDrag-Diff W/O DPT. & 38.63  & \textbf{0.895} \\
        StableDrag-Diff W/O CMS.  & 37.87 & 0.875 \\
        StableDrag-Diff & \textbf{36.36} & 0.893  \\
        \bottomrule
    \end{tabular}
    }
    \label{tab_ablation}
\vspace{-1mm}
\end{table}

\paragraph{\textbf{Confident motion supervision.}}
Here we study the effect of our confident motion supervision component. Firstly, we conduct experiments of the horse editing based on StableDrag-GAN. It can be seen from Fig.~\ref{fig:ablation1} that, as the confidence score gradually decreases, StableDrag without the confident motion supervision module produces low-quality editing image.
This indicates the importance of performing confident supervision at each step, and also demonstrates that the tracking score can reflect the quality of motion supervision.
As shown in Table~\ref{tab_ablation} ,the image fidelity decrease by 0.018 when substituting the confident motion supervision with original supervision method in DragGAN, which further substantiates the above conclusion.

\begin{table}[pt]
\begin{center}
\caption{Analysis on time consuming of training tracker and the drag process. Evaluation is performed on StableDrag-Diff.}
\fontsize{8.5}{9.5}\selectfont  
\setlength{\tabcolsep}{1.2mm}{
    \begin{tabular}{ccc|cc}  
    \toprule  
     Model& Tracker iters. & Drag steps & Tracker time(s) & Drag time(s) \cr  
    \midrule
    StableDrag-Diff & 1000 & 60 & 1.17 & 29.06 \cr
    StableDrag-Diff & 1000 & 80 & 1.08 & 38.80 \cr
    \bottomrule
    \end{tabular}
    }
\label{tab_time}
\vspace{-6mm}
\end{center}
\end{table}

\paragraph{\textbf{Discriminative point tracking.}}
In Fig.~\ref{fig:ablation2} and Table~\ref{tab_ablation}, we evaluate our StableDrag and the one without the discriminative tracking model.
We can see that StableDrag without the discriminative tracking model may suffer from misleading by the background distractor points, causing inaccurate results.
Especially, StableDrag-Diff without our discriminative tracking model increases the StableDrag-Diff by 2.27 of Mean Distance.
From the results, we can derive that the proposed discriminative tracking model helps the dragging model to achieve more accurate point-based image editing.

\paragraph{\textbf{Practicality of the tracking module.}}
The proposed point tracker is concise in both formulation and implementation. As shown in Table.~\ref{tab_time}, the training process of the tracker (about only 1 second) costs far less time than the drag process. 
As for the point tracking before each supervision step, it runs very fast since only a convolution operation should be performed.
It is worth noting that, during the point tracking process, we use a local search strategy to avoid discerning two completely similar objects (e.g., two almost identical dog) in global area.
Besides, the core code implementation is simple and easy to adapt to other related methods, since only around 60-rows code is added to the baseline. And we will release the code. 

\paragraph{\textbf{Sensitivity analysis on $\tau$ and $\lambda$.}}
To better understand the robustness of the proposed method, we have conducted sensitivity analysis on $\tau$ and $\lambda$ as in Table~\ref{tab-sensi_tau} and Table~\ref{tab-sensi_lam}.
Through the results, we can arrive that, \romannumeral1) the confident motion supervision is critical for stable dragging and a proper threshold is important, \romannumeral2) merging the proposed tracker with the original feature difference can obtain optimal dragging performance.

\begin{table}[pt]
    \centering
    \caption{\textbf{Sensitivity analysis on $\tau$}, where $\lambda$ is fixed to 0.0.}
    \vspace{-2mm}
    \label{tab-sensi_tau}
    \fontsize{8}{9}\selectfont
    \begin{tabular}{lcccccc}
        \toprule
        $\tau$ & 0.0 & 0.2 & 0.4 & 0.6 & 0.8 & 1.0 \\
        \midrule
        MD/IF & 42.1/0.868 & 41.6/0.874 & \textbf{39.8}/0.891 & 43.3/0.913 & 47.4/0.939 & 51.2/\textbf{0.955} \\
        \bottomrule
    \end{tabular}
\end{table}

\begin{table}[pt]
    \centering
    \caption{\textbf{Sensitivity analysis on $\lambda$}, where $\tau$ is fixed to 0.0.}
    \vspace{-2mm}
    \fontsize{8}{9}\selectfont
    \begin{tabular}{lcccccc}
        \toprule
        $\lambda$ & 0.0 & 0.2 & 0.4 & 0.6 & 0.8 & 1.0 \\
        \midrule
        MD/IF & 42.1/0.868 & 41.6/0.869 & 41.6/0.87 & \textbf{37.9}/\textbf{0.875} & 40.7/0.874 & 39.0/0.875 \\
        \bottomrule
    \end{tabular}
    \label{tab-sensi_lam}
\end{table}

\begin{figure}[!pt]
\centering
\includegraphics[width=\linewidth]{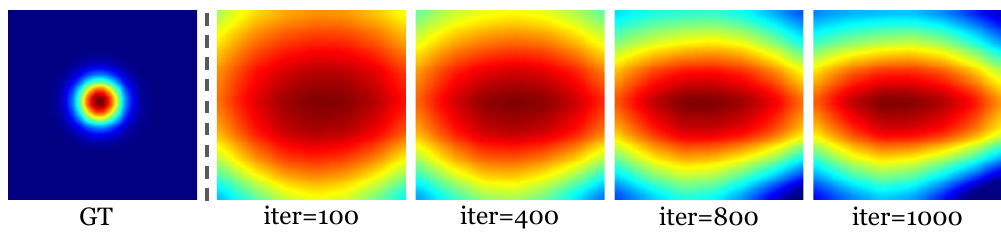}
\vspace{-6mm}
\caption{Visualization of the learning process for the tracking model $z_i$.}
\vspace{-6mm}
\label{fig:vis_track}
\end{figure}

\subsection{Visualization of learning process for $z_i$}
To give a more comprehensive understanding for the discriminative tracking model, in Fig.~\ref{fig:vis_track}, we visualize the prediction results of the tracking model during the learning process. It can be seen that, with the training iterations increasing, background points (i.e., points away from the center) are gradually suppressed, resulting in a more robust and discriminative point tracking model, which can help the dragging model to generate more accurate editing results.

%% file: sec/conclusion.tex
\section{Conclusion}
We have built a stable drag-based editing framework, coined as StableDrag, by designing a discirminative point tracking method and a confidence-based latent enhancement strategy for motion supervision.
With the proposed point tracking method, we can precisely locate the updated handle points, thereby boosting the stability of long-range manipulation. While the latter can guarantee the optimized latent as high-quality as possible across all the manipulation steps.
Thanks to the unique designs, we have instantiated two types of models including StableDrag-GAN and StableDrag-Diff to demonstrate the generality. Through extensive qualitative and quantitative experiments on a variety of examples, StableDrag has attained stable and precise drag performance. We expect our findings and analysis can facilitate the development of precise image editing.